\documentclass[runningheads]{llncs}
\usepackage{amsmath}
\usepackage{amssymb}
\usepackage{bm}
\usepackage{color}
\usepackage[dvipsnames]{xcolor}
\usepackage{graphicx}
\usepackage{mathbbol}
\usepackage{multirow, hhline}

\newcommand{\cth}{\multicolumn{1}{c}}
\newcommand{\cthe}{\multicolumn{1}{c|}}
\newcommand{\cthee}{\multicolumn{1}{c||}}

\newcommand{\ul}{\underline}
\newcommand{\figcaption}[1]{\def\@captype{figure}\caption{#1}}
\newcommand{\tblcaption}[1]{\def\@captype{table}\caption{#1}}

\newcommand{\sectionv}[1]{\vspace{-2mm}\section{#1}\vspace{-1mm}}
\newcommand{\subsectionv}[1]{\vspace{-3mm}\subsection{#1}\vspace{-2mm}}
\newcommand{\ryoshiha}[1]{\textcolor{red}{#1}}
\newcommand{\nonpr}[1]{\textcolor{red}{#1}} % 英文校正に出した後修正、慎重に文法などチェックする
\newcommand{\kurdyla}[1]{\textcolor{green}{#1}} %英文校正で修正された
\newcommand{\camera}[1]{\textcolor{red}{#1}} 

\renewcommand{\ryoshiha}[1]{\textcolor{black}{#1}}
\renewcommand{\nonpr}[1]{\textcolor{black}{#1}} % 英文校正に出した後修正、慎重に文法などチェックする
\renewcommand{\kurdyla}[1]{\textcolor{black}{#1}} %英文校正で修正された
\renewcommand{\camera}[1]{\textcolor{black}{#1}} 

\definecolor{watermarkcolor}{gray}{0.85}
\begin{document}
\title{Context-Free TextSpotter for Real-Time and Mobile End-to-End Text Detection and Recognition}
\author{Ryota Yoshihashi \and Tomohiro Tanaka \and Kenji Doi \and \\ Takumi Fujino \and Naoaki Yamashita}
\institute{Yahoo Japan Corporation\\
\email{{ryoshiha@yahoo-corp.jp}}}
\maketitle              % typeset the header of the contribution
\vspace{-6mm}
\begin{abstract}
In \kurdyla{the deployment of} scene-text \ryoshiha{spotting} systems on mobile platforms,
\kurdyla{lightweight} models with low computation are preferable.
In concept, end-to-end (E2E) text \ryoshiha{spotting} \nonpr{is} suitable for such purposes because \nonpr{it performs} text detection and recognition in a single model.
However, current state-of-the-art E2E methods rely on 
heavy feature extractors, recurrent sequence modellings, and complex shape aligners to pursue accuracy, \kurdyla{which means} their computations \kurdyla{are} still heavy.
We explore the opposite direction: \textit{{How far can \kurdyla{we} go without bells and whistles in E2E text \ryoshiha{spotting}?}} \kurdyla{To this end}, we propose a text-spotting method that consists of simple convolutions and \ryoshiha{a} few \kurdyla{post-processes}, named \textit{{Context-Free TextSpotter}}.
%Our experiments show that Context-Free TextSpotter achieves real-time E2E text recognition on GPUs with comparable accuracy to the state-of-the-art methods' for focused scene texts, and loss of accuracy is not large even for harder incidental scene texts. 
\ryoshiha{Experiments \kurdyla{using} standard benchmarks show that Context-Free TextSpotter achieves real-time text spotting on a GPU with only three million parameters, which is the smallest and fastest among existing deep text spotters, with an acceptable transcription quality degradation compared to \nonpr{heavier ones}.}
Further, we demonstrate that our text spotter \kurdyla{can} run on a smartphone
with affordable latency, \kurdyla{which} is valuable for building stand-alone OCR \camera{applications}.
\vspace{-2mm}
\end{abstract}
\vspace{-4mm}
\sectionv{Introduction}\vspace{-2mm}
\begin{figure}[t]
\centering
\includegraphics[width=.86\textwidth]{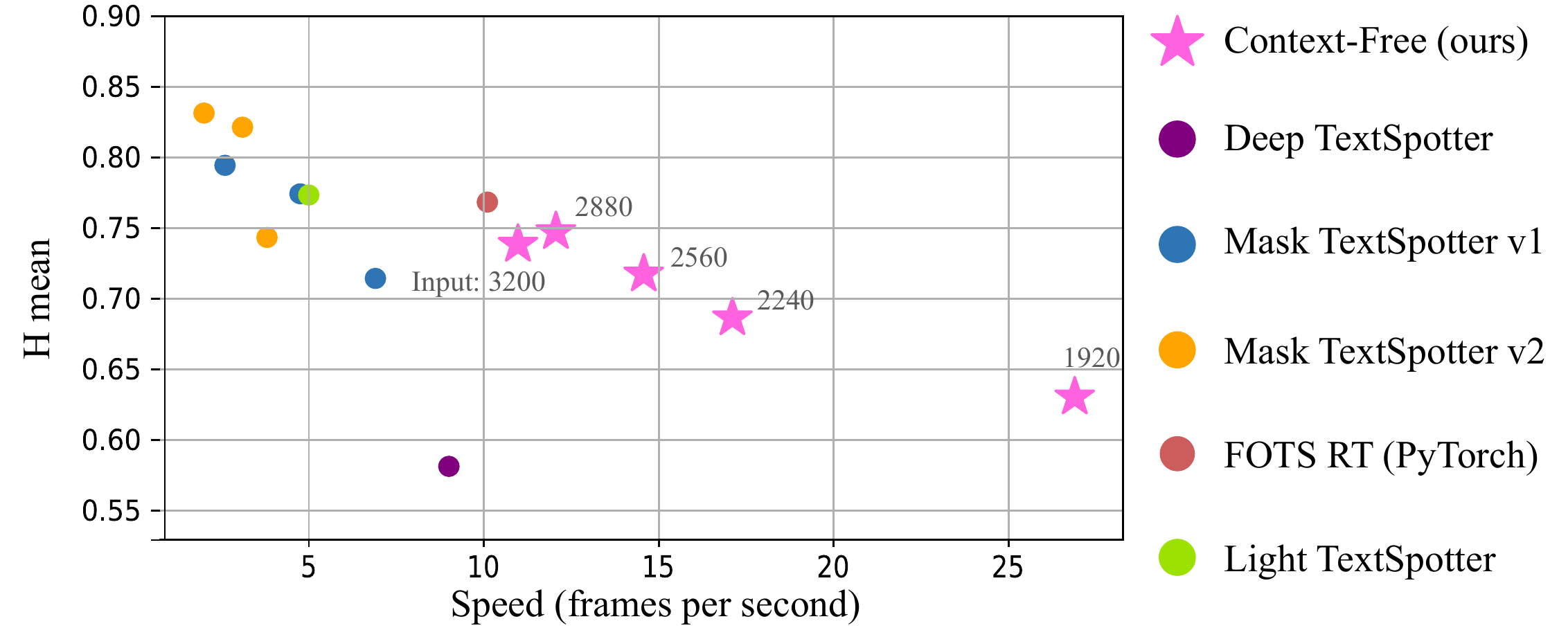}
\vspace{-4mm}
\caption{Recognition quality vs. speed in scene text spotters, evaluated \kurdyla{with a GPU} \kurdyla{on} ICDAR2015 incidental scene text benchmark with Strong lexicon.
\kurdyla{The proposed} Context-Free TextSpotter runs faster than existing ones with similar accuracy, and enables near-real-time text spotting with a certain accuracy degradation.} \label{fig:tradeoff}
\vspace{-6mm}
\end{figure}

\ryoshiha{Scene text spotting is a fundamental \kurdyla{task} with \kurdyla{a variety of} applications such as image-base translation, life \kurdyla{logging}, or industrial automation.}
To make the useful applications \ryoshiha{more conveniently accessible}, deploying text-spotting systems
on mobile devices \kurdyla{has shown promise}, as we \kurdyla{can} see \kurdyla{in the} recent spread of smartphones.
If text spotting runs client-side on mobile devices, users can enjoy the advantages of edge computing 
such as availability outside of the communication range, 
saving of packet-communication fees consumed by uploading images, and \kurdyla{fewer} concerns about privacy violation by leakage of uploaded images.
%Thus, text recognition methods that can be deployed in more types of devices will be much beneficial.

However, \kurdyla{while} recent deep text-detection and recognition methods have become highly accurate, \kurdyla{they are now} so heavy 
that they can not be easily deployed on mobile devices.
A {\it heavy} model may mean that its inference is computationally expensive, or that it has many parameters and its file size is large, but either is unfavorable for mobile users.
Mobile CPUs and GPUs are generally powerless \kurdyla{compared to the} ones \kurdyla{on} servers, so \kurdyla{the} inference latency of slow models may \kurdyla{become} \kurdyla{intolerable}.
\kurdyla{The large} file sizes of models are also problematic, because
they consume \kurdyla{a large amount of} memory and traffic\kurdyla{, both of which} are \kurdyla{typically} limited \kurdyla{on} smartphones.
For example, \kurdyla{the} Google Play app store has \kurdyla{limited the} maximum app size \kurdyla{to} 100MB so that developers \kurdyla{are cognizant of} this problem, but apps that \kurdyla{are equipped with} large deep recognition models may \kurdyla{easily} exceed the limit without careful compression.

The abovementioned problems motivated us to develop a lightweight text spotting method for mobiles.
For this purpose, \kurdyla{the} recently \kurdyla{popular} end-to-end (E2E) text spotting \kurdyla{shows promise}.
E2E recognition, which performs joint learning of modules (e.g., text detection and recognition) on a shared feature representation, has \kurdyla{been increasingly used in} deep learning to enhance \kurdyla{the} efficiency and effectiveness of models, and already much research effort has been dedicated to E2E text spotters~\cite{busta2017deep,liu2018fots,he2018end,liao2019mask,liao2020mask}. 
Nevertheless, most of \kurdyla{the} existing E2E text spotters are not speed-oriented; they typically run around five to eight frames per second (FPS) on high-end GPUs, which is not \kurdyla{feasible} for weaker mobile computation power.
The reasons for this heaviness \kurdyla{lies} in the complex design of modern text spotters.
\kurdyla{Typically}, a deep text spotter consists of 1) backbone networks as a feature extractor, 2) text-detection heads with RoI \kurdyla{pooling}, and 3) text-recognition heads.  
Each of these \kurdyla{components} has potential bottlenecks of computation; existing text spotters 1) adopt heavy backbones such as VGG-16 or ResNet50 \cite{busta2017deep},
2) often utilize RoI transformers customized for text recognition to geometrically align curved texts \cite{liu2018fots,liu2020abcnet},
and 3) the text-recognition head, which runs per text box, has unshared convolution and recurrent layers, which are \kurdyla{necessary for the} sequence modelling of word spells \cite{liu2018fots,he2018end}.

Then, how can \kurdyla{we make} an E2E text spotter light enough for mobile usage?
We argue that \kurdyla{the answer is} an E2E text spotter with as few additional 
modules to convolution as possible.
Since convolution is the simplest building block of deep visual systems 
and the de facto standard operation for which well-optimized kernels are offered in many environments, it has an intrinsic advantage in efficiency. 
Here, we propose {\it Context-Free TextSpotter}, 
an E2E text spotting method without bells and whistles.
It simply consists of lightweight convolution layers and a few \kurdyla{post-processes} to extract text polygons and character labels from the convolutional features.
This text spotter is {\it \kurdyla{context-free}} in \kurdyla{the} sense that it does not rely on \kurdyla{the} linguistic context conventionally provided by per-word-box sequence modelling \kurdyla{through} LSTMs, \kurdyla{or the} spatial context provided by geometric RoI transformations and geometry-aware pooling modules.
These \kurdyla{simplifications} may not seem straightforward, \kurdyla{as} text spotters \kurdyla{usually} needs sequence-to-sequence mapping to predict arbitrary-length texts from visual features.
Our idea to alleviate this complexity is to decompose a text spotter into a character spotter and a text-box detector that work in parallel.
In our framework, a character spotting module spots points in convolutional features that are readable as a character and \kurdyla{then} classifies the point features to give character labels to the points. Later text boxes and point-wise character labels are merged by a simple rule to construct word-recognition results.
Intuitively, characters are less curved in scenes than whole words are, and character recognition is easier to tackle without geometric operations. Sequence modelling becomes unnecessary in character classification, in contrast to \kurdyla{the} box-to-word recognition \kurdyla{utilized} in other methods.
Further, by introducing weakly supervised learning, \kurdyla{our method does not need} character-level annotations of real images to train the character-spotting part in our method. 

In experiments, we found that Context-Free TextSpotter worked surprisingly well
for its simplicity on standard benchmarks of scene text spotting.
It \kurdyla{achieved a} word-spotting Hmean \kurdyla{of} 84 with 25 FPS \kurdyla{on the} ICDAR 2013focused text dataset, 
and \kurdyla{an Hmean of} 74 with 12 FPS \kurdyla{on the} ICDAR2015 incidental text dataset \ryoshiha{in the Strong-lexicon protocol.}
Compared to existing text spotters, it is around three times faster than typical recent text spotters, while its recognition degradation is around five to ten percent-points.
\ryoshiha{Another advantage is \kurdyla{the} flexibility of the model: it can control the accuracy-speed balance simply \kurdyla{by} scaling input-image resolution within a single trained model, as shown in Fig.~\ref{fig:tradeoff}.}
Context-Free TextSpotter ran the fastest among deep text spotters
that reported their inference speed, and thus it is useful to extrapolate the current accuracy-speed trade-off curve into untrodden speed-oriented areas.
\nonpr{Finally, we demonstrate that our text spotter \kurdyla{can} run on a smartphone
with affordable latency, \kurdyla{which} is valuable for building stand-alone OCR applications.}

Our contributions summarize as follows: 1) we design a novel simple text-spotting framework called Context-Free TextSpotter. 2) We develop techniques useful for enhancing efficiency and effectiveness of our text spotter, including
the linear point-wise character decoding and the hybrid approach for character spotting, which are described in Sec. 3. 3) In experiments, we confirm that our text spotter runs the fastest among existing deep text spotters, and able to be deploy in iPhone to run with acceptable latency.

\sectionv{Related work}
\begin{figure}[t]
\centering
\includegraphics[width=0.92\textwidth]{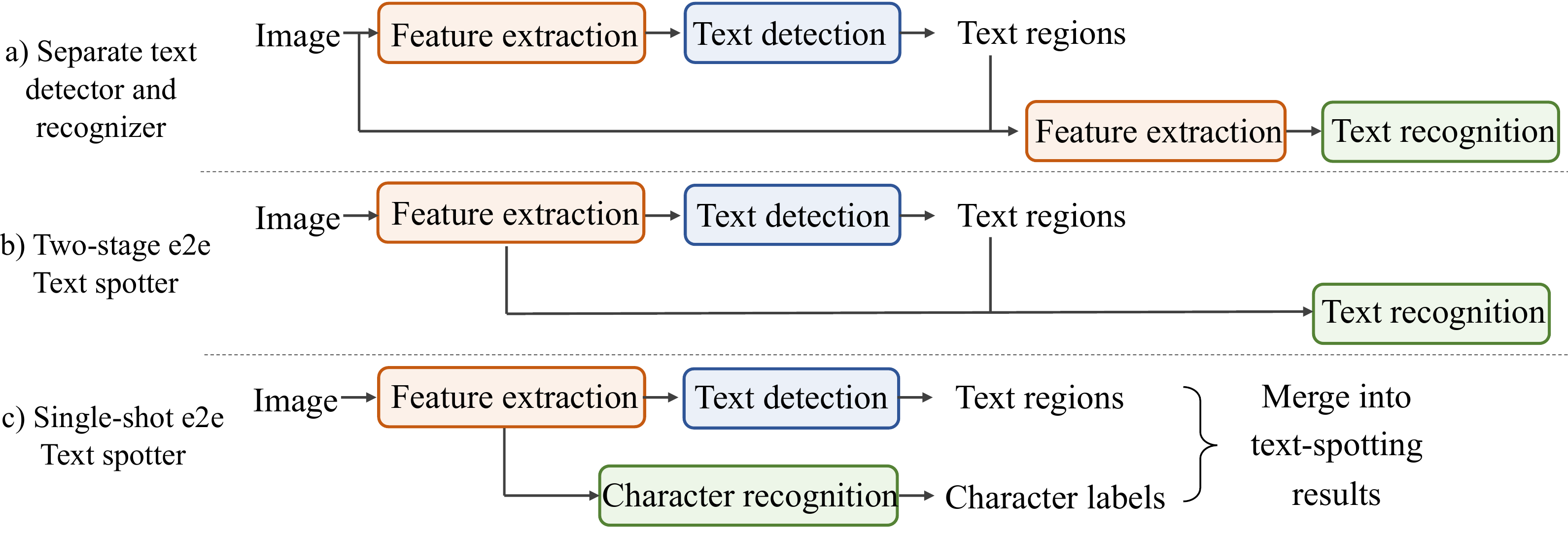}
\vspace{-4mm}
%\caption{The three typical text-spotting pipelines. a) Separate text detectors and recognizers: here the detector and recognizer share no computations and the recognizer extracts text-region features again from input images, which may mean redundant computation. b) Two-stage E2E text spotters reuse text-detection features for text recognition, which removes the twice feature extractions. However, the recognizer is still dependent on the detector's output regions, which results in less parallelizability.}
\caption{\kurdyla{Three} typical text-spotting pipelines. a) Non-E2E separate text detectors and recognizers, b) two-stage E2E text spotters, and c) single-shot text spotters. We use c) for the highest parallelizability.}
\vspace{-4mm}
\label{fig:concept}
\end{figure}
\subsectionv{Text detection and recognition}
Conventionally, text detection and recognition are regarded as related but separate tasks,
and \kurdyla{most} studies \kurdyla{have} focused on either one \kurdyla{or the other}.
%As a technical term {\it text recognition} refers to a task to transcribe well-cropped text images in narrow sense, while it may simply refer to a wider research area. 
%Classical methods for text recognition are often train a character classifiers on cropped letters and apply them to images in sliding-window manner~\cite{wang2012end}.
In text  recognition, while classical methods use character classifiers in a sliding-window manner~\cite{wang2012end},
more modern plain methods consist of convolutional feature extractors 
and recurrent-network-based text decoders~\cite{shi2016end}.
%Following are challenges and solutions in text recognition. 
%To predict arbitrary-length texts from visual feature, recurrent layers are widely used. 
Among recurrent architecture, bidirectional long short-term memories (BiLSTM~\cite{graves2005bidirectional}) is the most popular choice.
More recently, self-attention-based transformers, which \kurdyla{have been} successful in NLP, are intensively examined~\cite{zhu2019text}.
Connectionist temporal classification (CTC)~\cite{graves2006connectionist} is used as an objective function in sequence prediction, 
and later attention-based text decoding \kurdyla{has been} intensively researched as a more powerful alternative~\cite{cheng2017focusing}.
For not-well-cropped \kurdyla{or} curved texts, learning-based image rectification \kurdyla{has been} proposed~\cite{shi2016robust}.
These techniques are also useful in text spotting to design recognition branches.
%To boost text recognition accuracy, geometric transformation modules or attention modules are often added.

Deep text detectors are roughly categorized into two \kurdyla{types}: box-based detectors
and segment-based ones.
Box-based detectors \kurdyla{typically} estimate text \kurdyla{box} coordinates by regression from learned features. \nonpr{Such} techniques \kurdyla{have been} intensively studied in object detection
and they \kurdyla{are easily generalized} to text detection.
\kurdyla{However}, the box-based detectors often have difficulty in localizing curved texts accurately, unless any curve handling mechanism is adopted.
%\kurdyla{However}, the box-based detectors often have difficulty in localizing curved texts accurately \kurdyla{because} boxes cannot fit tightly on curved texts, unless any curve handling mechanism is adopted.
%There \kurdyla{are} many variants of box-based text detectors. 
TextBoxes~\cite{liao2018textboxes++} is directly inspired by an object detector SSD~\cite{liu2016ssd}, which regress rectangles from feature maps.
SSTD~\cite{he2017single} exploits text attention modules that are supervised by text-mask annotations.
EAST~\cite{zhou2017east} treats oriented texts by utilizing regression for box angles. 
CTPN~\cite{tian2016detecting} and RTN~\cite{zhu2017deep} similarly utilize strip-shaped boxes and connect them to represent curved text regions.

Segment-based detectors estimate foreground masks in a pixel-labeling manner, and extract their connected components as text instances.
\kurdyla{While} they are naturally able to handle oriented and curved texts, 
they sometimes mis-connect multiple text lines when the lines are \kurdyla{close together}.
For example, \kurdyla{fully convolutional} networks~\cite{long2015fully} can be applied to text detection with text/non-text mask prediction and some post processing of center-line detection and orientation estimation~\cite{zhang2016multi}.
PixelLink~\cite{deng2018pixellink} enhances the separability of text masks by introducing \kurdyla{8-neighbor pixel-wise} connectivity prediction in addition to text/non-text masks.
TextSnake~\cite{long2018textsnake} adopts \kurdyla{a} disk-chain representation that is predicted by text-line masks and radius maps.
CRAFT~\cite{baek2019character} enhances learning effectiveness by modeling character-region awareness and between-character affinity using a sophisticated supervisory-mask generation scheme, \nonpr{which we also adopt in our method}.

\vspace{-1mm}\subsectionv{End-to-end text recognition}
%End-to-end (E2E) learning refers to a class of learning methods that can link inputs to outputs with single learnable pipelines.
%In general, E2E methods can be more effective by optimizing their whole pipelines with regards to the final training objective than stage-wise learning,
%where each stage (e.g., detection and recognition) pursue their own objectives 
%disregarding to the other stages, and is not guaranteed to be optimal to the final objective.
%E2E methods also can be more efficient by sharing computations that multiple stages need
%(e.g., feature extraction for detection and recognition) than stage-wise methods, 
%which often need separate feature extractors for each stage.
End-to-end (E2E) learning refers to a class of learning methods that can link inputs to outputs with single learnable pipelines.
%, which is generally useful \kurdyla{for improving} learning efficiency and effectiveness.
\nonpr{It} has been successfully applied \kurdyla{for text recognition and detection}.
%Interestingly, the first E2E text recognition study ~\cite{wang2011end}, 
%which combined character detection by \kurdyla{Random Fern} and word construction \kurdyla{through a} pictorial structure, predates the boom of deep learning ~\cite{krizhevsky2012imagenet}.
%However, after \kurdyla{that pioneering} work
While the earliest method used a sliding-window character detector and word construction \kurdyla{through a} pictorial structure~\cite{wang2011end}, 
most of \kurdyla{the} modern E2E text recognizers \kurdyla{have been} based on deep CNNs. 
The \kurdyla{majority} of existing E2E text spotters are based on two-stage framework (Fig. \ref{fig:concept} b). Deep TextSpotter~\cite{busta2017deep} is the first deep E2E text spotter, that \kurdyla{utilizes} YOLOv2 as a region-proposal network (RPN) and cascades recognition branches of convolution + CTC after it.
FOTS~\cite{liu2018fots} adopts a similar framework of RPN and recognition branches, but improved \kurdyla{the} oriented text recognition by \kurdyla{adding an} RoI Rotate operation. 
Mask TextSpotter~\cite{lyu2018mask,liao2019mask,liao2020mask} has \kurdyla{an} instance-mask-based recognition head, that does not rely on CTC. 
\kurdyla{The recently proposed} CRAFTS~\cite{baek2020character} is a segment-based E2E text spotter \kurdyla{featuring enhanced recognition} by sharing character attention with \kurdyla{a} CRAFT-based detection branch.

Single-shot text spotting (Fig. \ref{fig:concept} c), in contrast, \kurdyla{has not been} \kurdyla{extensively} researched \kurdyla{up to} now.
CharNet~\cite{xing2019convolutional} utilizes parallel text-box detection and character semantic segmentation. Despite of its conceptual simplicity, it is not faster than two-stage spotters due to its heavy backbone and dense character labeling.
MANGO~\cite{qiao2021mango} exploits dense mask-based attention instead to eliminate \kurdyla{the} necessity for RoI operation. However, its character decoder relies on attention-based BiLSTM, and iterative inference is still needed, which might \kurdyla{create a} computational bottleneck in long texts. 
Here, our contribution is to show \kurdyla{that} a single-shot spotter \kurdyla{actually} runs faster than existing two-stage ones if proper simplification is done.

\vspace{-2mm}\subsectionv{Real-time image recognition}
%Outside of the document analysis field, we briefly review general ideas to speed up image recognition systems.
The large computation cost of deep nets is \kurdyla{an issue} in many fields, and
\kurdyla{various studies have explored} general techniques \kurdyla{for reducing} computation cost, \kurdyla{while others focus on} designing lightweight models for a specific tasks.
Some of examples of the former are pruning~\cite{han2015learning}, quantization~\cite{courbariaux2015binaryconnect,hubara2017quantized}, and lighter operators~\cite{han2020ghostnet,chen2020addernet}.
While \kurdyla{these} are \kurdyla{applicable to general models}, it is not clear whether the loss of computational precision they cause is within a tolerable range in text localization, which \kurdyla{require} high exactness for correct reading.
For the latter, lightweight models are seen in many vision tasks such as image classification, object detection, \kurdyla{and} segmentation.
However, lightweight text spotting is much less studied, seemingly due to the complexity of the task, although there \kurdyla{has been} some \kurdyla{research on} mobile text detection~\cite{fu2019text,decker1a2020mobtext,cordova2019pelee,jeon2020compact}. We are aware of only one \camera{previous} work: Light TextSpotter~\cite{guan2020light}, \kurdyla{which} is a slimmed \kurdyla{down} mask-based text spotter that features distillation and \kurdyla{a} ShuffleNet backbone.
Although it is 40\%-lighter than Mask TextSpotter, its inference speed is not as fast as ours due to \kurdyla{its} big mask-based recognition branch.
%For the latter, the most spotlighted success of real-time recognition is perhaps in generic object detection recently. The vision community investigates high-power two-stage methods and fast single-shot methods in parallel to broadening feasible applications.

\vspace{-2mm}\sectionv{Method}\vspace{-2mm}
\begin{figure}[t]
\includegraphics[width=.94\textwidth]{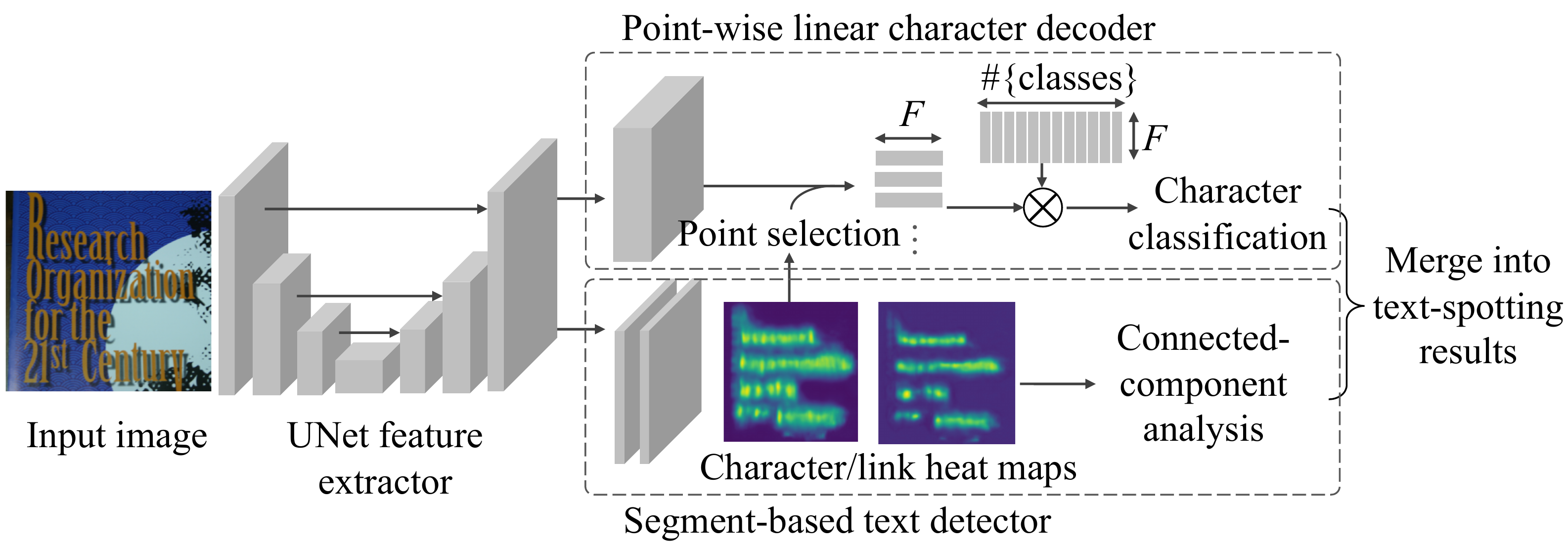}
\vspace{-4mm}
\caption{\kurdyla{O}verview of proposed Context-Free TextSpotter.} \label{fig:overview}
\vspace{-5mm}
\end{figure}

The design \kurdyla{objective} behind the proposed Context-Free Text Spotter is to pursue minimalism
in text spotting.
As a principle, deep E2E text spotters need to be able to detect text boxes in images and recognize their contents, while the detector and recognizer can share feature representations they exploit.
Thus, a text spotter, at \kurdyla{minimum}, needs to have a feature extractor, a text-box detector, and a text recognizer. 

\kurdyla{With text spotters}, the text recognizers \kurdyla{tend to} become the most complex \kurdyla{part of the system}, \kurdyla{as} 
text recognition from detected boxes \kurdyla{needs} to perform a sequence-to-sequence prediction,
that is to \kurdyla{relates} an arbitrary-length feature sequence to an arbitrary-length text.
Also, the \kurdyla{recognizer's} computation \kurdyla{depends} on \kurdyla{the} detector's output, which makes the pipeline less parallelizable (Fig.~\ref{fig:concept} b).
%To \kurdyla{get rid of} sequence-to-sequence prediction, 
We break down the text recognizer into a character spotter and classifier, \kurdyla{where} the character spotter pinpoints coordinates of characters within uncropped feature maps of whole images, and the character classifier classifies each of the spotted characters regardless \kurdyla{of} other characters around it or the text box \kurdyla{to which} it belongs (Fig.~\ref{fig:concept} c). 

The overview of Context-Free Text Spotter is shown in Fig.~\ref{fig:overview}.
It consists of 1) a U-Net-based feature extractor, 2) heat-map-based character and text-box detectors, and 
3) a character decoder \kurdyla{called} \emph{the point-wise linear decoder},
\kurdyla{which is} specially tailored for our method.
The later part of this section describes the details of each module.

\subsectionv{Text-box detector and character spotter}\label{sec:cspot}
\kurdyla{To take advantage of shape flexibility} and interpretability, we adopt segment-based methods
\kurdyla{for the} character and text-box detection.
More specifically, we roughly follow \kurdyla{the} CRAFT~\cite{baek2019character} text detector in \kurdyla{the} text-box localization procedure.
\kurdyla{Briefly}, our text-box detector generates two heat maps\kurdyla{: a region map and an affinity map}.
The region map is trained to activate strongly around the centers of characters, and  
the affinity map is trained to activate in areas between characters within single text boxes. In inference, connected components in the sum of the region map and affinity map are extracted as text instances.

For character spotting, we reuse the region map of the CRAFT-based text detector, and thus we do not need to add extra modules to our network.
Further, we adopt point-based character spotting, 
instead of \kurdyla{the} more common box- or polygon-based character detection,
to eliminate \kurdyla{the} necessity for box/polygon regression and RoI pooling operations.

\begin{figure}[t]
\centering
\includegraphics[width=.95\textwidth]{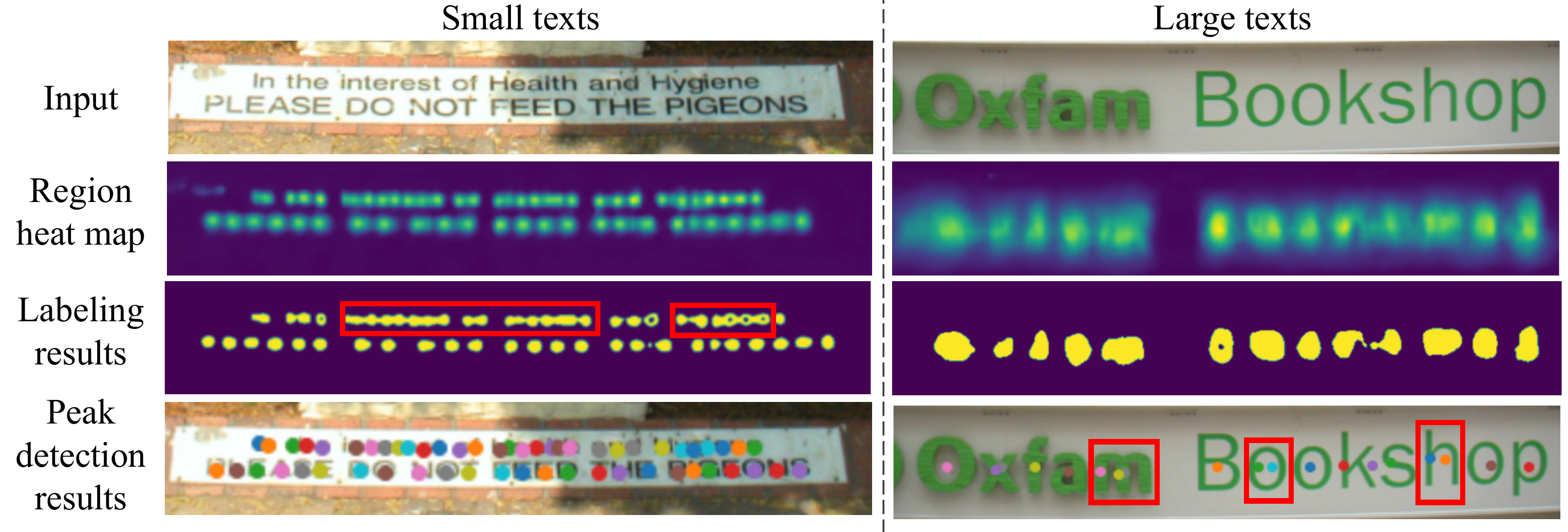}
\vspace{-4mm}
\caption{
%The scale matters in character spotting. 
Examples of small and large texts. Corresponding region heat maps (second row) show different characteristics 
that lead labeling-based approaches (third row) to failure (red boxes) in small texts, and peak-detection-based approaches (last row) to failure in large texts.
%The labels tend to fail to disconnect close character in small texts, 
%and the peaks tend to cause duplicated detection in large texts, as shown in red boxes.
%Following this observation, 
Thus, we adopt \kurdyla{a} hybrid approach in character spotting.}
%as described in Sec.~\ref{sec:cspot}.} 
\label{fig:scale}
\vspace{-5mm}
\end{figure}

To select character points from the region map, we prefer simple image-processing-based techniques to learning-based ones in order to avoid unnecessary overheads.
We consider two approaches: a labelling-based approach that performs grouping of heat maps, and \kurdyla{a} peak-detection based approach that picks up local maxima of the heat map. The former is adopted for spotting large characters, and the latter for spotting small characters.
Here, our insight helpful to the point selector \kurdyla{design} is that the form of heat maps' activation to characters largely differs by the scales of the characters in input images.
Examples of heat-map activation to large and small characters are shown in Fig.~\ref{fig:scale}.
For small characters, labels tend to fail to disconnect close characters, while they cause less duplicated detection of single characters.
Apparently, such tendency comes from the difference of activation patterns in region heat maps; the heat-map activation to large texts reflects detailed shapes of characters \kurdyla{that} may cause multiple peaks in them, while that to small texts \kurdyla{results in} simpler single-peaked blobs.

For labelling-based spotting for large characters, we use connected-component analysis~\cite{wu2009optimizing} to link separate characters. First we binarize region heat maps by a threshold, then link the foregrounds, and finally pick up centroids of each connected component.
For peak-based spotting for small objects, we use local maxima of region heat maps, namely
\vspace{-3mm}\begin{equation}
P = \left\{(x, y) \mid R(x, y) - \max_{(dx, dy) \in \mathcal{N}}\left(R(x + dx, y + dy)\right) = 0 \right\},
\vspace{-3mm}\end{equation}
where $R$ denotes the region heat map and $\mathcal{N}$ denotes 8-neighbors in the 2-D coordinate system.
\nonpr{Later, the extracted points from the labelling-based spotter are used in character decoding if they are included in {\it large} text boxes, and the ones from peak-based spotting are used if included in {\it small} text boxes, where {\it large} and {\it small} are defined by whether the shorter side of the box is longer or shorter than a certain threshold.} 

\subsectionv{Character decoder}
Our character decoder is called {\it \kurdyla{the} linear point-wise decoder}, 
which simply performs linear classification of feature vectors on detected character points.
Given a set of points $P = [(x_1, y_1), (x_2, y_2), ... , (x_N, y_N)]$, a feature map $\bm{f}$ with $F$ channels and size $H \times W$, and the number of classes (i.e., types of characters) $C$, the decoder is denoted as 
\vspace{-2mm}\begin{eqnarray}
    \text{cls} &=& \text{Softmax}(\bm{f}[:, P]\bm{w} - \mathbb{1}\bm{b}), \label{eqn:cls} \\ \nonumber
        &=& \text{Softmax}\left(\left(\begin{array}{c}
             \bm{f}[:, y_1, x_1] \\
             \bm{f}[:, y_2, x_2] \\
             \vdots \\
             \bm{f}[:, y_N, x_N] \\
        \end{array}\right)
        \left(\begin{array}{cccc} w^1_1 & w^2_1 & \dots & w^C_1 \\ w^1_2 & w^2_2 & \dots & w^C_2 \\ \vdots & \vdots & \dots & \vdots \\ w^1_F & w^2_F & \dots & w^C_F \end{array}\right) -  
        \left(\begin{array}{cccc} b_1 & b_2 &\dots& b_C\\ b_1 & b_2 &\dots& b_C \\ \vdots & \vdots & \dots & \vdots \\ b_1 & b_2 &\dots& b_C \end{array}\right) \right),
\vspace{-3mm}\end{eqnarray}
where $f[:,P]$ denotes the index operation that extracts feature vectors at the points $P$ and stacks them into \kurdyla{an} $(N, F)$-shaped matrix.
The parameters of linear transformation $(\bm{w}, \bm{b})$ are parts of \kurdyla{the} learnable \camera{parameters} of the model, and \kurdyla{are} optimized by the back
propagation during training jointly to the other network parameters,
where $\bm{w}$ is an $F \times C$ matrix and $\bm{b} = [b_1, b_2, \dots, b_C]$ is a $C$-dimensional vector \kurdyla{broadcast} by $\mathbb{1} = [1, 1, \dots, 1]^T$ with length $N$.
Finally, row-wise softmax is taken along the channel axis to form classification probabilities.
Then $\text{cls}$ becomes an $(N, C)$-shaped matrix, where its element at $(i, j)$ encodes the probability that the $i$-th points is recognized as the $j$-th character.  

After giving character probability to each point, we further filter out the points that are not confidently classified by applying a threshold.
\camera{Finally, each character point is assigned to the text box that include it, and for each text box, the character points that are assigned to it are sorted by x coordinates and read left-to-right.}

\begin{figure}[t]
\centering
\includegraphics[width=1.0\textwidth]{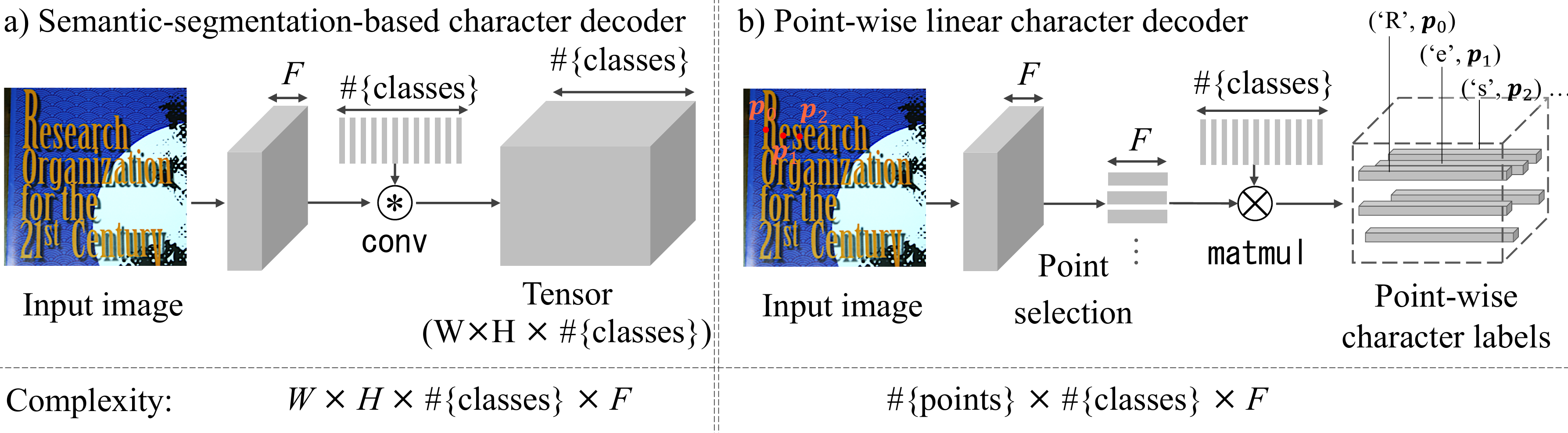}
\vspace{-4mm}
\caption{Comparison between semantic-segmentation-based character decoding and our linear point-wise decoding. The linear point-wise decoding has advantages in space and time complexity when \#\{points\} is small and \#\{classes\} is large.}
\vspace{-5mm}
\label{fig:complexity}
\end{figure}

It is worth noting that the linear point-wise decoding is numerically equivalent to
semantic-segmentation-style character decoding, except \kurdyla{that it is} computed sparsely. By regarding the classification weights and biases in Eq.~\ref{eqn:cls} as a $1 \times 1$ convolution layer, it can be applied
densely to the feature map, as shown in Fig.~\ref{fig:complexity} a.
%While such implementation would be attractive for its \kurdyla{fully convolutional} property,
\nonpr{Despite of its implementation simplicity,}
it suffers from its heavy output tensor.
The output has a $(W \times H \times \text{\#\{classes\}})$-sized label space,
where $W \times H$ is the desired output resolution and $\text{\#\{classes\}}$ is
the number of types of characters we want to recognize.
In text recognition, $\text{\#\{classes\}}$ is already large in \kurdyla{the} Latin script, which has over 94 \nonpr{alphanumerics and symbols} in total, and may be even larger in other scripts (for example, \kurdyla{there are} over 2,000 types of Chinese characters). 
In \kurdyla{practice}, dense labeling of 94 characters from a \nonpr{$(400\times400\times32)$-sized} feature map  consumes around 60 MB memory and 0.48 G floating-point operations
only by the output layer, regardless \kurdyla{of} how many letters are present in the image.
If given 2,000 characters, the output layer needs 1.2 GB memory, which critically limits \kurdyla{the} training and inference efficiency.
Thus, we prefer the point-wise alternative (Fig.~\ref{fig:complexity} b) to maintain scalability and to \kurdyla{avoid} unnecessary computation when \kurdyla{there are} few letters in images.

\subsectionv{Feature extractor}
\kurdyla{M}ost scene text spotters use VGG16 or ResNet50 
as their backbone feature extractors, \nonpr{which are heavy for mobiles}.
Our choice for \kurdyla{lightweight} design is CSP-PeleeNet~\cite{wang2020cspnet}, 
\kurdyla{which} was originally designed for mobile object detection~\cite{wang2018pelee},
and further lightened by adding cross-stage partial connection~\cite{wang2020cspnet} without reducing ImageNet accuracy.
The network follows \kurdyla{the} DenseNet~\cite{huang2017densely} architecture but \kurdyla{is} substantially smaller
\kurdyla{thanks to} having 32, 128, 256, 704 channels in its 1/1-, 1/4-, 1/8-, 1/16-scaled stage each, and a total \kurdyla{of} 21 stacked dense blocks.

On the CSPPeleeNet backbone, we construct U-Net~\cite{ronneberger2015u} by adding upsampling modules for \camera{feature decoding}.
We put the output layer of the U-Net at the 1/4-scaled stage, which corresponds to the second block in the CSPPeleeNet. 
\footnote{Note that Fig.~\ref{fig:overview} does not \kurdyla{precisely} reflect the structure due to layout constraints.}

\vspace{-2mm}\subsectionv{Training}\vspace{-0mm}
Context-Free TextSpotter is end-to-end differentiable and trainable with ordinary gradient-based solvers.
The training objective $L$ is defined as
\vspace{-3mm}\begin{eqnarray}
L &=& L_{\text{det}} + \alpha L_{\text{rec}}, \\
L_{\text{det}} &=& \frac{1}{WH}\left(\left|R - R_{\text{gt}}\right|^2 + \left|A - A_{\text{gt}}\right|^2\right), \\
L_{\text{rec}} &=& \frac{1}{N} \sum_{(x_{\text{gt}}, y_{\text{gt}}, c_{\text{gt}}) \in P_{\text{gt}}} \text{CE}\left(\text{Softmax}\left(\bm{f}[:, y_{\text{gt}}, x_{\text{gt}}]\bm{w} - \bm{b}\right), c_{\text{gt}}\right),
\vspace{-0mm}\end{eqnarray}\vspace{-0mm}
where $(W, H)$ denotes feature-map size, $R, A, R_{\text{gt}}, A_{\text{gt}}$ denote the predicted region and affinity map and their corresponding ground \kurdyla{truths}, and $(x_{\text{gt}}, y_{\text{gt}}, c_{\text{gt}}) \in P_{\text{gt}}$ denotes character-level ground truths that indicate locations and classes of the characters. CE denotes the cross-entropy classification loss, and $\alpha$ is a hyperparameter that balances the two \kurdyla{losses}.

\camera{This training objective needs character-level annotations, but many scene text recognition benchmarks only provide word-level ones.
Therefore, we adopt weakly-supervised learning that exploits
approximated character-level labels in the earlier training stage, 
and updates the labels by self-labeling in the later stage.}
In \kurdyla{the} earlier stage, for stability, we fix $P_{gt}$ to the centers of {\it approximated character polygons}.
Here, the approximated character polygons are calculated by equally dividing the word polygons into the word-length parts along their word lines.
The ground truths of \kurdyla{the} region and affinity maps are generated using the same approximated character polygons following the same manner \kurdyla{as the} CRAFT~\cite{baek2019character} detector. 

In \kurdyla{the} later stage, $P_{\text{gt}}$ is updated by \kurdyla{self-labeling:} the word regions of training images are cropped by ground-truth polygon annotation, the network under training predicts heat maps from the cropped images, and spotted points in them are used as new $P_{\text{gt}}$ combined with the ground-truth word transcriptions. Word instances whose number of spotted points and length of the ground-truth transcription are not equal are \kurdyla{discarded} because the difference suggests inaccurate \kurdyla{self-labeling}.
We also exploit synthetic text images with character-level annotation by setting their $P_{\text{gt}}$ at the character-polygon centers.

\vspace{-2mm}\sectionv{Experiments}\vspace{-1mm}
To evaluate \kurdyla{the} effectiveness and efficiency of Context-Free TextSpotter, we conducted intensive experiments \kurdyla{using} scene-text recognition datasets.
%First, we compared our method \kurdyla{with} existing text spotting and detection methods in \kurdyla{terms} of detection and recognition quality, model sizes, and inference speed.
Further, we analyzed factors that control quality and speed, namely, choice of backbones \kurdyla{and} input-image resolution\nonpr{, and conducted ablative analyses of the modules}.
Finally, we deployed Context-Free TextSpotter as an iPhone application and measured \kurdyla{the} on-device inference speed to demonstrate the feasibility of mobile text spotting.

\vspace{-3mm}\paragraph{Datasets}
We used ICDAR2013~\cite{karatzas2013icdar} and ICDAR2015~\cite{karatzas2015icdar} dataset for evaluation. 
%\kurdyla{These datasets} were derived from Robust Reading Competition (RRC). 
ICDAR2013 thematizes focused scene texts and consists of 229 training and 233 testing images. ICDAR2015 thematizes incidental scene texts and consists of 1,000 training and 500 testing images.
In \kurdyla{the} evaluation, we followed the competition's official protocol. %using the provided evaluation script.
\kurdyla{For} detection, \camera{we used DetEval for ICDAR2013 and IoU with threshold 0.5 for ICDAR2015}, and calculated precision, recall, and their harmonic means (H means).
\kurdyla{For} recognition, we used {\it end-to-end} and {\it word spotting} protocol, with provided {\it Strong} (100 words), {\it Weak} (1,000 words), and {\it Generic} (90,000 words) lexicons.

\kurdyla{For} training, we additionally used \kurdyla{the} SynthText~\cite{gupta2016synthetic} and ICDAR2019 MLT \cite{nayef2019icdar2019} datasets to supplement the relatively small training sets of ICDAR2013/2015.
SynthText is a synthetic text dataset that provides 800K images and character-level annotations by polygons. 
ICDAR2019 MLT is a multi-lingual dataset with 10,000 training images. We replaced non-Latin scripts \kurdyla{with} {\it ignore} symbols \kurdyla{so that we could use} the dataset to train \kurdyla{a} Latin text spotter.

\begin{table}[t]
\addtolength{\tabcolsep}{1mm}
\caption{Text detection and recognition results \kurdyla{on} ICDAR2015. {\bf Bold} \kurdyla{indicates} the best \kurdyla{performance} and \ul{underline} \kurdyla{indicates} the second best within \kurdyla{lightweight} \ryoshiha{text-spotting} models. * \kurdyla{indicates} our re-measured or re-calculated numbers, while the others are excerpted from the literature.}\label{tab:ic15}
\vspace{-3mm}
\scriptsize
%\fontsize{7pt}{0pt}\selectfont
\begin{tabular}{|l||p{5mm}p{5mm}p{5mm}||p{8mm}p{8mm}p{8mm}|p{8mm}p{8mm}p{8mm}||p{4mm}p{4mm}|}
\hline
\multirow{2}{*}{Method} & \multicolumn{3}{|c||}{Detection} & \multicolumn{3}{|c|}{Word spotting} & \multicolumn{3}{|c||}{End-to-end} & Prms & \multirow{2}{*}{FPS} \\ \cline{2-10}
 & \cth{P} & \cth{R} & \cthee{H} & \cth{S} & \cth{W} & \cthe{G} & \cth{S} & \cth{W} & \cthee{G} & (M) &  \\ \hline  \hline
\emph{Standard models} &\cth{} & \cth{} & \cthee{} & \cth{} & \cth{} & \cthe{} & \cth{} & \cth{} & \cthee{} &  &  \\
EAST~\cite{zhou2017east} &\cth{83.5} & \cth{73.4} & \cthee{78.2} & \cth{--} & \cth{--} & \cthe{--} & \cth{--} & \cth{--} & \cthee{--} & -- & 13.2 \\
Seglink~\cite{shi2017detecting} &\cth{73.1} & \cth{76.8} & \cthee{75.0} & \cth{--} & \cth{--} & \cthe{--} & \cth{--} & \cth{--} & \cthee{--} & -- & 8.9 \\
CRAFT~\cite{baek2019character} &\cth{89.8} & \cth{84.3} & \cthee{86.9} & \cth{--} & \cth{--} & \cthe{--} & \cth{--} & \cth{--} & \cthee{--} & 20* & 5.1* \\
StradVision~\cite{karatzas2015icdar} &\cth{--} & \cth{--} & \cthee{--} & \cth{45.8} & \cth{--} & \cthe{--} & \cth{43.7} & \cth{--} & \cthee{--} & -- & -- \\
Deep TS~\cite{busta2017deep} &\cth{--} & \cth{--} & \cthee{--} & \cth{58} & \cth{53} & \cthe{ 51} & \cth{54} & \cth{51} & \cthee{47} & -- & 9 \\
Mask TS~\cite{he2018end} &\cth{91.6} & \cth{81.0} & \cthee{86.0} & \cth{79.3} & \cth{74.5} & \cthe{64.2} & \cth{79.3} & \cth{73.0} & \cthee{62.4} & 87* & 2.6 \\
FOTS~\cite{liu2018fots} &\cth{91.0} & \cth{85.1} & \cthee{87.9} & \cth{84.6} & \cth{79.3} & \cthe{63.2} & \cth{81.0} & \cth{75.9} & \cthee{ 60.8} & 34 & 7.5 \\
Parallel Det-Rec~\cite{li2019new} &\cth{83.7} & \cth{96.1} & \cthee{89.5} & \cth{89.0} & \cth{84.4} & \cthe{68.8} & \cth{85.3 } & \cth{80.6} & \cthee{65.8} & -- & 3.7 \\
CharNet~\cite{xing2019convolutional} &\cth{89.9} & \cth{91.9} & \cthee{90.9} & \cth{--} & \cth{--} & \cthe{--} & \cth{83.1} & \cth{79.1} & \cthee{69.1} & 89 & 0.9* \\
CRAFTS~\cite{baek2020character} &\cth{89.0} & \cth{85.3} & \cthee{87.1} & \cth{--} & \cth{--} & \cthe{--} & \cth{83.1} & \cth{82.1} & \cthee{74.9} & -- & 5.4 \\
\hline

\emph{Lightweight models} &\cth{} & \cth{} & \cthee{} & \cth{} & \cth{} & \cthe{} & \cth{} & \cth{} & \cthee{} &  &  \\
PeleeText~\cite{cordova2019pelee} &\cth{85.1} & \cth{72.3} & \cthee{78.2} & \cth{--} & \cth{--} & \cthe{--} & \cth{--} & \cth{--} & \cthee{--} & 10.3 & 11 \\
PeleeText++~\cite{cordova2020pelee} &\cth{81.6} & \cth{\ul{78.2}} & \cthee{79.9} & \cth{--} & \cth{--} & \cthe{--} & \cth{--} & \cth{--} & \cthee{--} & 7.0 & 15 \\
PeleeText++ MS~\cite{cordova2020pelee} &\cth{87.5} & \cth{76.6} & \cthee{81.7} & \cth{--} & \cth{--} & \cthe{--} & \cth{--} & \cth{--} & \cthee{--} & 7.0 & 3.6 \\
Mask TS mini~\cite{he2018end} &\cth{--} & \cth{--} & \cthee{--} & \cth{71.6} & \cth{63.9} & \cthe{51.6} & \cth{\ul{71.3}} & \cth{62.5} & \cthee{50.0} & 87* & 6.9 \\
FOTS RT~\cite{liu2018fots} &\cth{85.9} & \cth{\bf{79.8}} & \cthee{\ul{82.7}} & \cth{\ul{76.7}} & \cth{\ul{69.2}} & \cthe{53.5} & \cth{\bf{73.4}} & \cth{\bf{66.3}} & \cthee{\ul{51.4}} & 28 & 10* \\
Light TS~\cite{guan2020light} &\cth{\bf{94.5}} & \cth{70.7} & \cthee{80.0} & \cth{\bf{77.2}} & \cth{\bf{70.9}} & \cthe{\bf{65.2}} & \cth{--} & \cth{--} & \cthee{--} & 34* & 4.8 \\
Context-Free (ours)&\cth{\ul{88.4}} & \cth{77.1} & \cthee{\bf{82.9}} & \cth{74.3} & \cth{67.1} & \cthe{\ul{54.6}} & \cth{70.2} & \cth{\ul{63.4}} & \cthee{\bf{52.4}} & \bf{3.1}  & \bf{12} \\
\hline
\end{tabular}
\vspace{-2mm}
\end{table}
\begin{table}[t]
\addtolength{\tabcolsep}{1mm}
\caption{Text detection and recognition results \kurdyla{on} ICDAR2013.}\label{tab:ic13}
\vspace{-3mm}
\scriptsize
\begin{tabular}{|l||p{5mm}p{5mm}p{5mm}||p{8mm}p{8mm}p{8mm}|p{8mm}p{8mm}p{8mm}||p{7mm}p{5mm}|}
\hline
\multirow{2}{*}{Method} & \multicolumn{3}{|c||}{Detection} & \multicolumn{3}{|c|}{Word spotting} & \multicolumn{3}{|c||}{End-to-end} & Params & \multirow{2}{*}{FPS} \\ \cline{2-10}
 & \cth{P} & \cth{R} & \cthee{H} & \cth{S} & \cth{W} & \cthe{G} & \cth{S} & \cth{W} & \cthee{G} & \cth{(M)} &  \\ \hline  \hline
\emph{Standard models} &\cth{} & \cth{} & \cthee{} & \cth{} & \cth{} & \cthe{} & \cth{} & \cth{} & \cthee{} &  &  \\
EAST~\cite{zhou2017east} &\cth{92.6} & \cth{82.6} & \cthee{87.3} & \cth{--} & \cth{--} & \cthe{--} & \cth{--} & \cth{--} & \cthee{--} & -- & 13.2 \\
Seglink~\cite{shi2017detecting} &\cth{87.7} & \cth{83.0} & \cthee{85.3} & \cth{--} & \cth{--} & \cthe{--} & \cth{--} & \cth{--} & \cthee{--} & -- & 20.6 \\
CRAFT~\cite{baek2019character} &\cth{97.4} & \cth{93.1} & \cthee{95.2} & \cth{--} & \cth{--} & \cthe{--} & \cth{--} & \cth{--} & \cthee{--} & 20* & 10.6* \\
StradVision~\cite{karatzas2015icdar} &\cth{--} & \cth{--} & \cthee{--} & \cth{85.8} & \cth{82.8} & \cthe{70.1} & \cth{81.2} & \cth{78.5} & \cthee{67.1} & -- & -- \\
Deep TS~\cite{busta2017deep} &\cth{--} & \cth{--} & \cthee{--} & \cth{92} & \cth{89} & \cthe{81} & \cth{89} & \cth{86} & \cthee{77} & -- & 10 \\
Mask TS~\cite{he2018end} &\cth{91.6} & \cth{81.0} & \cthee{86.0} & \cth{92.5} & \cth{92.0} & \cthe{88.2} & \cth{92.2} & \cth{91.1} & \cthee{86.5} & 87* & 4.8 \\
FOTS~\cite{liu2018fots} &\cth{--} & \cth{--} & \cthee{88.3} & \cth{92.7} & \cth{90.7} & \cthe{83.5} & \cth{88.8} & \cth{87.1} & \cthee{80.8} & 34 & 13* \\
Parallel Det-Rec~\cite{li2019new} &\cth{--} & \cth{--} & \cthee{--} & \cth{95.0} & \cth{93.7} & \cthe{88.7} & \cth{90.2} & \cth{88.9} & \cthee{84.5} & -- & -- \\
CRAFTS~\cite{baek2020character} &\cth{96.1} & \cth{90.9} & \cthee{93.4} & \cth{--} & \cth{--} & \cthe{--} & \cth{94.2} & \cth{93.8} & \cthee{92.2} & -- & 8.3 \\
\hline

\emph{Lightweight models} &\cth{} & \cth{} & \cthee{} & \cth{} & \cth{} & \cthe{} & \cth{} & \cth{} & \cthee{} &  &  \\
MobText~\cite{decker1a2020mobtext} &\cth{\ul{88.3}} & \cth{66.6} & \cthee{76.0} & \cth{--} & \cth{--} & \cthe{--} & \cth{--} & \cth{--} & \cthee{--} & 9.2 & -- \\
PeleeText~\cite{cordova2019pelee} &\cth{80.1} & \cth{79.8} & \cthee{80.0} & \cth{--} & \cth{--} & \cthe{--} & \cth{--} & \cth{--} & \cthee{--} & 10.3 & 18 \\
PeleeText++~\cite{cordova2020pelee} &\cth{87.0} & \cth{73.5} & \cthee{79.7} & \cth{--} & \cth{--} & \cthe{--} & \cth{--} & \cth{--} & \cthee{--} & 7.0 & 23 \\
PeleeText++ MS~\cite{cordova2020pelee} &\cth{\bf{92.4}} & \cth{\ul{80.0}} & \cthee{\bf{85.7}} & \cth{--} & \cth{--} & \cthe{--} & \cth{--} & \cth{--} & \cthee{--} & 7.0 & 3.6 \\
Context-Free (ours)&\cth{85.1} & \cth{\bf{83.4}} & \cthee{\ul{84.2}} & \cth{83.9} & \cth{80.0} & \cthe{69.1} & \cth{80.1} & \cth{76.4} & \cthee{67.1} & \bf{3.1} & \bf{25} \\
\hline
\end{tabular}
\vspace{-5mm}
\end{table}

\vspace{-3mm}\paragraph{Implementation details}
Character-recognition feature channels $F$ was set to 256.
Input images were resized into 2,880 pixels for ICDAR2015 and 1,280 pixels for ICDAR2013 in the longer side keeping the aspect ratio.
\kurdyla{For} training, we used Adam with initial learning rate 0.001, recognition-loss weight 0.01, and the batch size \kurdyla{of} five, where three \kurdyla{are} from \kurdyla{real the} and the rest \kurdyla{are} from SynthText training images.
\kurdyla{For} evaluation, we used weighted edit distance~\cite{lyu2018mask} for lexicon matching.
Our implementation was based on PyTorch-1.2.0 and ran on a virtual machine with an Intel Xeon Silver 4114 CPU, 16GB RAM, and an NVIDIA Tesla V100 (VRAM16GB$\times$2) GPU. All run times were measured with the batch size \camera{of one}.

\vspace{-2mm}\subsectionv{Results}
\begin{figure}[t]
\centering
\includegraphics[width=0.95\textwidth]{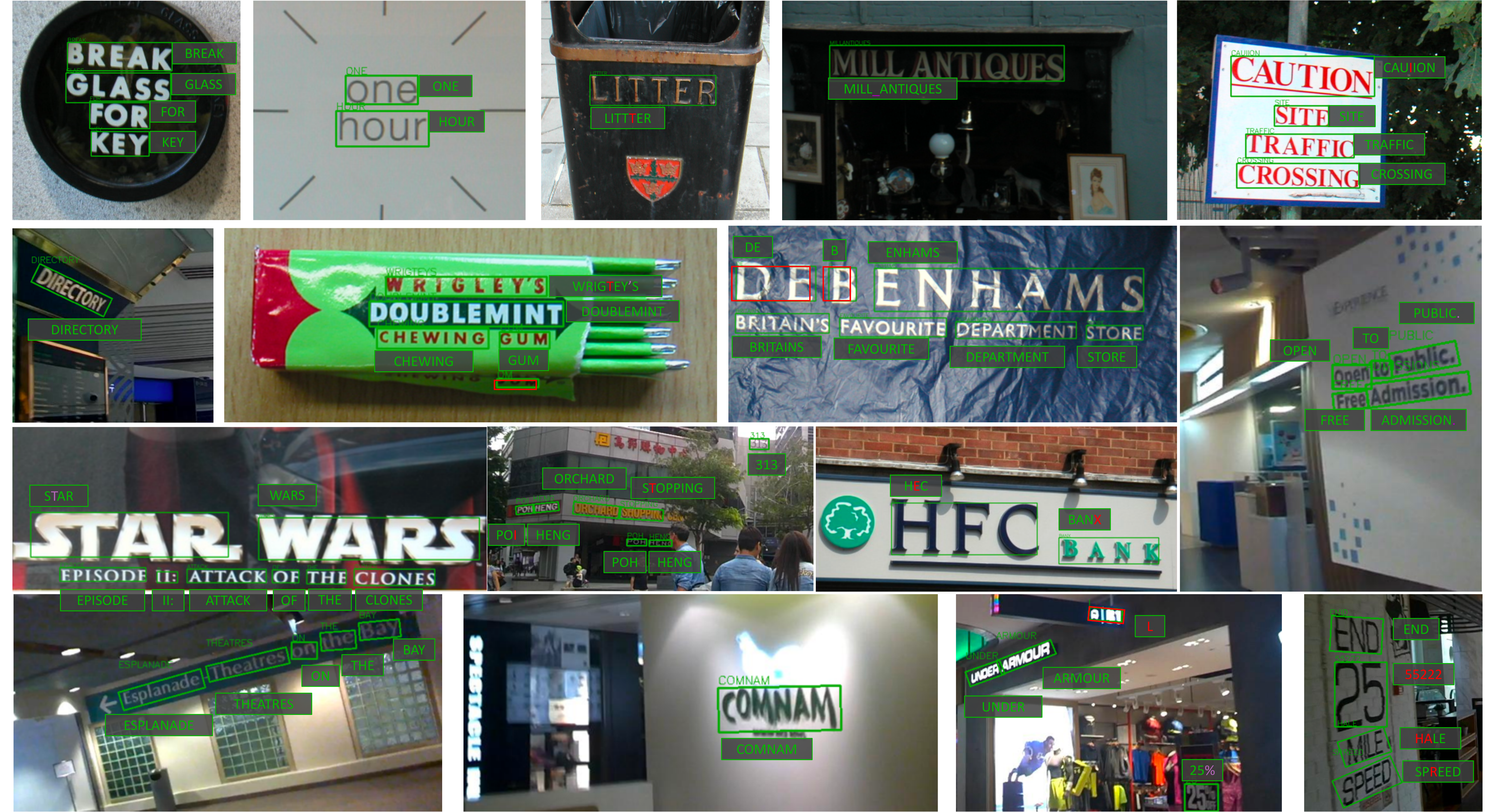}
\vspace{-4mm}
\caption{Text spotting results with Context-Free TextSpotter \kurdyla{on} ICDAR2013 and 2015. The lexicons are not used in this visualization. Red characters indicate wrong recognition and purple missed ones. Best viewed zoomed in digital.} \label{fig:examples}
\vspace{-6mm}
\end{figure}

Table~\ref{tab:ic15} summarizes the detection and recognition results \kurdyla{on} ICDAR2015. 
Context-Free TextSpotter \kurdyla{achieved an} 82.9 detection H mean and 74.3 word-spotting H mean with \kurdyla{the} Strong lexicon with 3.8 million parameters and 12 FPS \kurdyla{on} a GPU.
The inference speed and model size of Context-Free \kurdyla{was} the best among \kurdyla{all compared \ryoshiha{text spotters}}.

For \kurdyla{a} detailed comparison, we \kurdyla{separated} the text spotters \kurdyla{into} two \kurdyla{types}: {\it standard models} and {\it \kurdyla{lightweight} models}. 
While it is \kurdyla{difficult} to introduce such separation due to the continuous nature of the size-performance balance in neural networks, we \kurdyla{categorized} 1) models that are claimed to be \kurdyla{lightweight} in their papers and 2) smaller or faster versions reported in papers whose focus \kurdyla{were} creating  standard-sized models. 
\kurdyla{The} PeleeText~\cite{cordova2019pelee,cordova2020pelee} family is detection-only models based on Pelee~\cite{wang2018pelee}, but the model sizes are larger than ours due to feature-pyramid-based bounding box regression.
Mask TextSpotter mini (Mask TS mini) and FOTS Real Time (FOTS RT) are smaller versions reported in the corresponding literature, and are acquired by scaling input image size. However, their model sizes are large due to the full-size backbones, while the accuracies \kurdyla{were} comparable to ours. This suggests that simple minification of standard models is not optimal in repurposing for mobile or fast inference.
On another note, FOTS RT was originally reported to run in 22.6 FPS in a customized Caffe with a TITAN-Xp GPU.
Since it is closed-source, we re-measured a 3rd-party reimplementation in PyTorch\footnote{https://github.com/jiangxiluning/FOTS.PyTorch},
and the FPS was 10 when the batch size \kurdyla{was} one. We reported the latter number in the table in order to prioritize comparisons in the same environment.
\ryoshiha{The other methods that we re-measured showed faster speed than in their original reports, seemingly due to our newer GPU.}
Light TextSpotter, \kurdyla{which} is based on lightened Mask TextSpotter, \kurdyla{had an} advantage in \kurdyla{the} evaluation with \kurdyla{the} Generic lexicon, but is much heavier than ours due to the instance-mask branches in the network.

Table~\ref{tab:ic13} summarizes the detection and recognition results \kurdyla{on} ICDAR2013.
ICDAR2013 \kurdyla{contains} focused texts with relatively large and clear appearance, which enables good recognition accuracy in smaller test-time input-image sizes than in ICDAR2015.
While all text spotters \kurdyla{ran} faster, Context-Free \kurdyla{was} again the fastest among them.
Light TextSpotter and the smaller versions of FOTS and Mask TextSpotter were not experimented in ICDAR2013 and thus \kurdyla{are} omitted \kurdyla{here}. This \kurdyla{means} we have no \kurdyla{lightweight} competitors in text spotting, but ours outperformed \kurdyla{the} lightweight text detectors while our model size was smaller and our inference was faster than theirs. 
%\ryoshiha{We again confirmed that \kurdyla{our model} outperformed detection-only lightweight models in detection H mean and recall.}

 To analyze the effect of input-image sizes \kurdyla{on} inference, we collected the results with different input sizes \kurdyla{and visualize them in} Fig.~\ref{fig:tradeoff}. Here, input images were resized using bilinear interpolation keeping their aspect ratios before inference. We \kurdyla{used} a single model trained with cropped images $960\times960$ in size without retraining.
 The best results \kurdyla{were} obtained when the longer sides were 2,880 pixels, where the speed was 12 FPS. We also \kurdyla{obtained} \kurdyla{the} near-real-time speed of 26 FPS with inputs 1,600 pixels long, even \kurdyla{on the more difficult} ICDAR2015 dataset with 62.8 H mean.
 
Table~\ref{tab:ablation} summarizes the ablative analyses by removing \kurdyla{the} modules and techniques we adopted in the final model.
Our observations are four-fold: First, real-image training is critical (a vs. b1). Second, end-to-end training itself did not much improve detection (b1 vs. b2), but removing boxes that were not validly read as texts (i.e., ones transcribed as empty strings by our recognizer) \kurdyla{significantly} boosted the detection score (c1). Third, \kurdyla{the} combination of peak- and labeling-based character spotting (called hybrid) was useful (c1--3). Fourth, weighted edit distance~\cite{liao2019mask} and adding MLT training data~\cite{li2019new}, the known techniques in the literature, \kurdyla{were} also helpful for our \kurdyla{model} (d and e). 
Table~4 shows comparisons of the different backbones in implementing our method. We tested  MobileNetV3-Large~\cite{howard2019searching} and GhostNet~\cite{han2020ghostnet} as recent fast backbones,
and confirmed that CSPPeleeNet was the best among them.
 
%Table~4 shows comparisons of the different backbones in implementing our method.
%We additionally evaluated MobileNetV3-Large~\cite{howard2019searching} and GhostNet~\cite{han2020ghostnet} as fast backbones,
%and confirmed that CSPPeleeNet was the best among them.
%A possible problem in MobileNetV3 is that it excessively reduces the lower-stage feature channels,
%which we use in character classification. MobileNetV3 has only 24 channels in its 1/4-scaled stage.
%GhostNet was better than MobileNetV3 but did not match CSPPeleeNet, and it was the slowest among the tested backbones. While the numbers of floating-point operations of MobileNetV3 and GhostNet are similar, GhostNet has many copy operations in it, which could affect the inference speed.

Finally, as a potential drawback of context-free text recognition, we found that duplicated reading of a single characters and minor \camera{spelling errors} sometimes appeared in outputs uncorrected, as seen in examples in Fig.~\ref{fig:examples} and the recognition metrics with the {\it Generic} lexicon. These may become a problem more when the target lexicon is large.
Thus, we investigated the effect of lexicon sizes on text-spotting accuracy. We used {\it Weak} as a basic lexicon and inflated it by adding randomly sampled words from {\it Generic}. The trials were done five times and we plotted the average. 
The results are shown in Fig.~\ref{fig:lexicon}, which shows the recognition degradation was relatively gentle when the lexicon size is smaller than 10,000. This suggests that Context-Free TextSpotter is to some extent robust against large vocabulary, despite it lacks language modelling.

\begin{table}[t]
\addtolength{\tabcolsep}{1mm}
\caption{Ablative analyses of modules and techniques we adopted.}\label{tab:ablation}
%\scriptsize
\fontsize{8pt}{0.75pt}\selectfont
\centering
\vspace{-3mm}
\begin{tabular}{rccccc|cc}
   & Training & E2E & Removing  & Character  & Lexicon  & Det H & WS S \\ 
& &  &  unreadables &  spotting &  matching &  & \\  \hline
a) & Synth & & -- & -- & -- & 64.8 & -- \\
b1) & + IC15 & & -- & -- & -- & 79.9 & -- \\
b2) & + IC15 & \checkmark & & Peak & Edit dist. & 80.2 & -- \\
c1) & + IC15 & \checkmark & \checkmark & Peak & Edit dist. & 81.0 & 70.5 \\
c2) & + IC15 & \checkmark & \checkmark & Label & Edit dist. & 78.4 & 48.4  \\
c3) & + IC15 & \checkmark & \checkmark & Hybrid & Edit dist. & 81.0 & 71.3 \\
d) & + IC15 & \checkmark & \checkmark & Hybrid & Weighted ED & 81.0 & 72.8 \\
e) & + IC15 \& 19 & \checkmark & \checkmark & Hybrid & Weighted ED & 82.9 & 74.3 \\
\end{tabular}
\vspace{-1mm}
\end{table}

\begin{figure}[t]
  \begin{minipage}[c]{.45\textwidth}
    \addtolength{\tabcolsep}{1mm}
    {{\bf Table 4.} Comparisons of lightweight backbones in our framework.}\label{tab:backbone}\vspace{2mm}
    \centering
    %\vspace{5mm}
    \vspace{5mm}\begin{tabular}{c|cc|c}
    Backbone & IC13 & IC15 & FPS \\ \hline
    MobileNetV3 & 67.1 & 50.1 & 11 \\
    GhostNet & 74.6 & 63.3 & 9.7 \\
    CSPPeleeNet & 83.9 & 74.3 & 12 \\
    \end{tabular}
    %\vspace{-2mm}
  \end{minipage}
  \hfill
  \begin{minipage}[c]{.5\textwidth}
    \centering
    \vspace{-2mm}
    \includegraphics[width=0.94\textwidth]{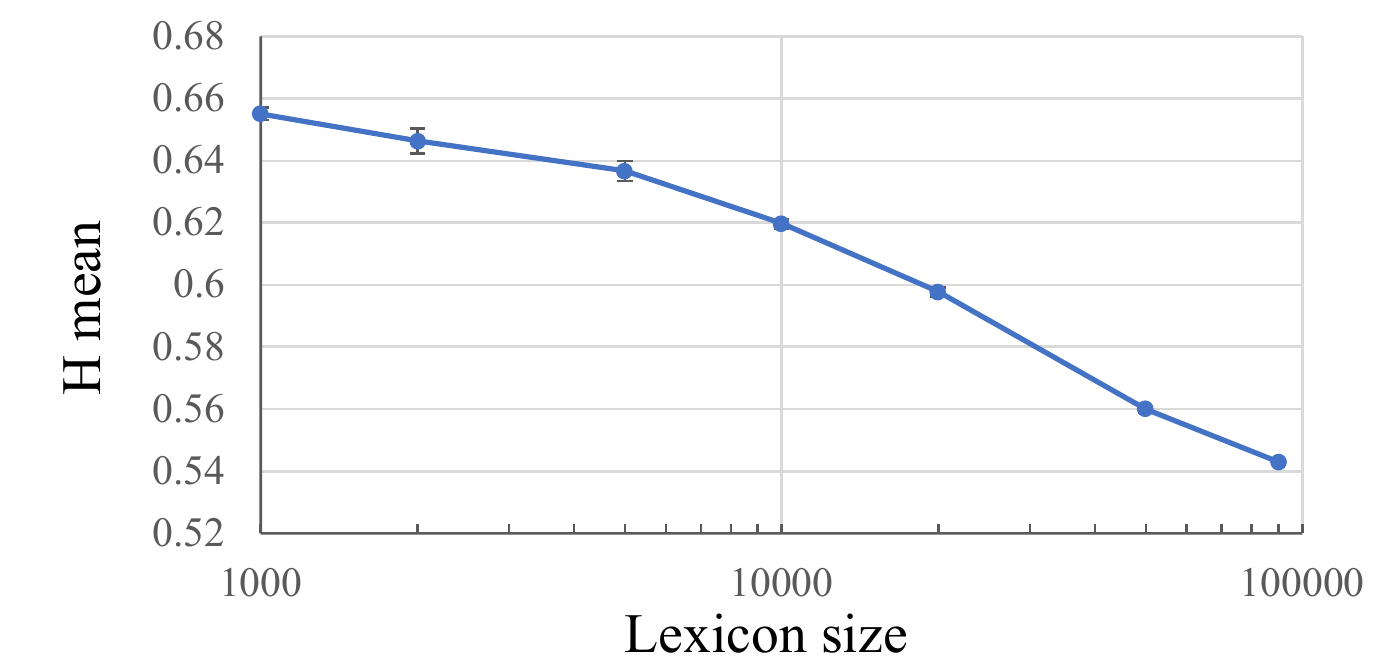}
    \vspace{-6mm}
    \caption{Lexicon size vs. recognition H mean.} \label{fig:lexicon}
    %\vspace{-4mm}
  \end{minipage}
  \vspace{-6mm}
\end{figure}

\begin{figure}[t]
\centering
\vspace{-4mm}
\includegraphics[width=0.92\textwidth]{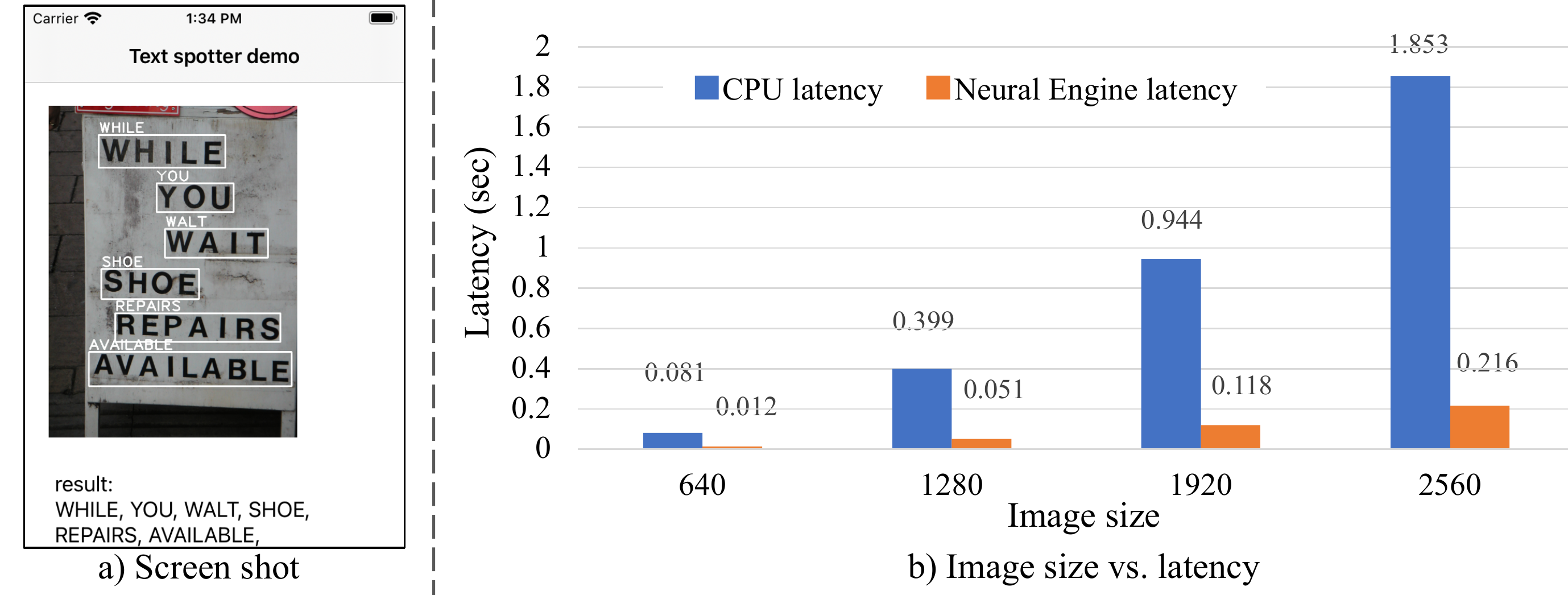}
\vspace{-4mm}
\caption{On-device Benchmarking with iPhone 11 Pro.} \label{fig:mobile}
\vspace{-4mm}
\end{figure}
\vspace{-2mm}
\subsectionv{On-device Benchmarking}
To confirm \kurdyla{the} feasibility of deploying our text spotter on mobiles, we conducted on-device benchmarking. We used \kurdyla{an} iPhone 11 Pro \camera{with the Apple A13 CPU and 4GB memory}, and ported our PyTorch model to \kurdyla{the} CoreML framework to run on it. We also implemented Swift-based post-processing (i.e., connected-component analysis and peak detection) to include their computation times in \kurdyla{the} benchmarking. With our ICDAR2013 setting (input size $= 1,280$), the average inference latency was 399 msec with \kurdyla{a} CPU, and 51 msec with \kurdyla{a} Neural Engine~\cite{neuralengine} hardware accelerator. Usability studies suggest that latency within 100 msec makes``the user feel that the system is reacting instantaneously'', and 1,000 msec is ``the limit for the user's flow of thought to stay uninterrupted, even though the user will notice the delay''~\cite{nielsen1994usability}. Our text spotter \kurdyla{achieves} the former if an accelerator is given, and the latter on a mobile CPU. \nonpr{More latencies with different input sizes are shown in Fig.~\ref{fig:mobile}}.

\camera{We also tried to port existing text spotters.
MaskTextSpotters \cite{liao2019mask,liao2020mask} could not be ported to CoreML, due to their customized layers. While implementing the special text-spotting layers in the CoreML side would solve this,
we also can say that simpler convolution-only models, like ours, have an advantage in portability.
Another convolution-only model, CharNet \cite{xing2019convolutional} ran on iPhone but fairly slowly.
With NeuralEngine, it took around 1.0 second to process 1,280-pixel-sized images, and could not process larger ones due to memory limits.
With the CPU, it took 8.1 seconds for 1,280-pixel images, 18.2 seconds for 1,920, and ran out of memory for larger sizes.}

\vspace{-2mm}\sectionv{Conclusion}\vspace{-2mm}
\kurdyla{In this work, we have proposed Context-Free TextSpotter},
an end-to-end text spotting method without bells and whistles \kurdyla{that enables real-time and mobile text spotting}.
We hope our method inspires \kurdyla{the} developers for OCR applications, and \kurdyla{will} serve as a \kurdyla{stepping} stone for researchers who want to prototype their ideas quickly. 

\vspace{-2mm}\section*{Acknowledgements}\vspace{-4mm}
We would like to thank Katsushi Yamashita, 
Daeju Kim, and members of the AI Strategy Office in SoftBank for helpful discussion.

\vspace{-4mm}
\bibliographystyle{splncs04}
%\tiny
%\bibliography{ocr}

\end{document}